\documentclass{article}
\usepackage{arxiv}
\usepackage[utf8]{inputenc}
\usepackage[T1]{fontenc}
\usepackage[english]{babel}
\usepackage{amsmath,amssymb,amsthm,dsfont}
\usepackage{graphicx,booktabs,array,placeins,float}
\usepackage{microtype}
\usepackage[numbers,sort&compress]{natbib}
\usepackage[colorlinks=true,allcolors=blue]{hyperref}
\usepackage{url}
\graphicspath{{figures/}}
\newtheorem{theorem}{Theorem}
\newtheorem{proposition}[theorem]{Proposition}
\newtheorem{corollary}[theorem]{Corollary}
\theoremstyle{remark}

\renewcommand{\shorttitle}{Prescribed-Parameter Convolutional Kernels}
\title{A Mechanistic Interpretation of Prescribed-Parameter Convolutional Kernels}
\author{Jorge Marco Blanco\\Universidad Carlos III de Madrid
\And Rub\'en Cuevas Rum\'in\\Universidad Carlos III de Madrid}
\date{}
\hypersetup{
 pdftitle={A Mechanistic Interpretation of Prescribed-Parameter Convolutional Kernels},
 pdfauthor={Jorge Marco Blanco and Ruben Cuevas Rumin},
 pdfsubject={Time series classification and prescribed-parameter convolutional representations},
 pdfkeywords={ROCKET, time series classification, PPV, random features, effective dimensionality}}
\begin{document}
\raggedbottom
\maketitle
\begin{abstract}
Prescribed-parameter convolutional transforms achieve strong time series classification performance using fixed kernels, simple pooling operations, and a linear classifier. Their success raises a mechanistic question: which properties of the representation make this construction effective? We study the proportion of positive values (PPV), a pooling operator shared by ROCKET, MiniROCKET, and MultiROCKET. We establish exact invariance to circular translations under periodic boundaries and quantify departures from invariance under finite boundaries. A complementary analysis bounds the sensitivity of a smooth PPV surrogate. We then define the population pairwise geometry induced by a distribution of convolutional parameters and show how a finite random kernel bank concentrates around that geometry. This separates the role of the parameter distribution from the number of sampled kernels. Experiments on the UCR classification collection examine effective dimensionality, distance correlation with the input, and classification accuracy using PPV-only, unit-energy ROCKET kernels. Increasing kernel count produces diminishing representational gains. Increasing kernel length generally raises effective dimensionality and distance correlation, while the best test performance occurs at intermediate lengths and training accuracy remains close to perfect. Together, these results support an interpretation in which nonlinear pooling creates a richer but structured representation that can favour linear separation, while further representational richness need not improve generalisation.
\end{abstract}
\section{Introduction}
\label{sec:int-introduction}
Some of the strongest approaches to time series classification depart from the usual assumption that useful convolutional representations must be learned. ROCKET applies a bank of random convolutional kernels, pools their responses into features, and trains a linear classifier~\cite{dempster2020rocket}. MiniROCKET replaces most of the randomness with a prescribed construction~\cite{dempster2021minirocket}; MultiROCKET enriches the pooling and input representations~\cite{tan2022multirocket}; and Hydra introduces competition between kernels~\cite{dempster2023hydra}. Their empirical performance suggests that useful class structure can be made accessible without learning the convolutional weights or fitting a complex classifier.

This observation leaves a mechanistic gap. Why can a large bank of prescribed filters followed by simple nonlinear pooling produce an effective representation? What does randomness contribute, and why do intermediate kernel lengths remain useful even when longer kernels yield richer representations? An answer should distinguish properties of the transform itself from choices made by the classifier fitted afterwards.

We use ROCKET as the experimental platform and focus on the proportion of positive values (PPV), retaining one feature per kernel and discarding the maximum response. PPV is retained by ROCKET, MiniROCKET, and MultiROCKET and contributes strongly to their classification performance~\cite{dempster2020rocket,dempster2021minirocket,tan2022multirocket}. Our empirical construction additionally normalises centred kernel weights to unit energy. This modification is stated explicitly because it changes the parameter distribution relative to the reference ROCKET implementation.

The analysis makes three contributions. First, we establish exact translation invariance of PPV features under periodic boundaries, quantify the influence of finite boundaries, and analyse the stability of a smooth surrogate. Second, we define the population pairwise geometry induced by random convolutional parameters and bound its approximation by a finite kernel bank. The parameter distribution determines the geometry, whereas the number of kernels controls the concentration of its empirical approximation. Third, we examine how kernel count and kernel length affect effective dimensionality, distance correlation, and classification performance. These interventions separate an increase in the number of sampled features from a change in the distribution that generates them.

The results support a geometric interpretation of the transform. Nonlinear PPV pooling distributes observations over additional effective directions while retaining statistical dependence with the input geometry. This is consistent with the intuition behind Cover's theorem~\cite{cover1965geometrical}, under which a nonlinear embedding can make class structure more amenable to linear separation. The interpretation also has a limit: longer kernels generally increase the geometric measures without producing a corresponding improvement in test performance. We discuss this behaviour as compatible with possible overfitting, rather than as a direct identification of its cause.

\section{Background and Related Work}
\label{sec:background}
\subsection{Time series, convolution, and linear classification}
\label{sec:conv1d}
A univariate time series is an ordered vector $\mathbf{x}\in\mathbb{R}^{N}$. A labelled dataset contains $M$ such series, with class labels in $\{1,\ldots,C\}$. The ordering of observations matters: discriminative information can lie in patterns occurring at different temporal positions and scales. A convolution applies the same kernel throughout the series, and dilation extends its temporal reach without increasing the number of weights.

For a kernel with weights $\mathbf{w}\in\mathbb{R}^{l}$, bias $b$, and positive integer dilation $d$, the valid response is
\begin{equation}
 u[n]=b+\sum_{j=1}^{l}w_j x[n+(j-1)d],
 \qquad n=1,\ldots,N-(l-1)d.
 \label{eq:conv}
\end{equation}
Its receptive field spans $(l-1)d+1$ input positions. We follow the deep-learning convention of calling this operation convolution, although the absence of weight reversal makes it a cross-correlation. Padding determines which response positions are retained and how the series is extended outside its observed interval. Pooling then summarises each response series into a fixed number of features. For a bank of $K$ kernels, the resulting feature vector lies in $\mathbb{R}^{D}$; in the PPV-only construction, $D=K$.

The prescribed-parameter family separates feature extraction from supervised fitting. The convolutional parameters are sampled or prescribed in advance, although some constructions calibrate biases from training data. Class labels are used to fit the readout. A ridge classifier is well suited to the resulting feature matrices because they often contain many correlated coordinates and more features than training observations~\cite{hoerl1970ridge,hastie2009elements}. For binary targets $\widetilde y^{(i)}\in\{-1,+1\}$ and feature vectors $\mathbf z^{(i)}$, it minimises
\begin{equation}
 \sum_{i=1}^{M}\bigl(\widetilde y^{(i)}-\boldsymbol\beta^\top\mathbf z^{(i)}-\beta_0\bigr)^2
 +\lambda\|\boldsymbol\beta\|_2^2,
 \qquad \lambda>0.
 \label{eq:ridge}
\end{equation}
Multiclass classification uses one-versus-rest scores. Regularisation stabilises estimation when features are correlated; it can be selected using training-data cross-validation.

\subsection{Prescribed-parameter convolutional transforms}
\label{sec:family}
\paragraph{ROCKET.}
ROCKET draws kernel lengths uniformly from $\{7,9,11\}$, samples Gaussian weights and mean-centres them, draws biases uniformly from $[-1,1]$, and varies dilation and padding~\cite{dempster2020rocket}. The dilation is sampled as
\begin{equation}
 d=\lfloor 2^a\rfloor,\qquad
 a\sim\mathcal U\left(0,\log_2\frac{N-1}{l-1}\right).
\end{equation}
Padding is chosen randomly between valid convolution and length-preserving zero padding. Each kernel contributes PPV and the maximum convolutional response. The default bank has $10{,}000$ kernels. Its design was established empirically: increasing kernel count improves classification with diminishing returns, while several alternative parameter choices perform comparably. Our analysis retains the sampling structure but uses PPV alone and the unit-energy modification described in Section~\ref{subsec:experimental_protocol}.

\paragraph{MiniROCKET.}
MiniROCKET uses 84 prescribed weight patterns of length 9, formed by placing three coefficients equal to $2$ among six coefficients equal to $-1$~\cite{dempster2021minirocket}. Dilation spans multiple temporal scales and biases are calibrated from training convolution outputs. It retains PPV alone and makes the transform almost deterministic. Its effectiveness shows that independently sampled Gaussian weights are not necessary for a useful convolutional representation.

\paragraph{MultiROCKET.}
MultiROCKET applies a related kernel construction to both the original series and its first-order difference~\cite{tan2022multirocket}. It complements PPV with the mean of positive values, mean of indices of positive values, and longest stretch of positive values. These operators capture response magnitude, location, and persistence in addition to frequency. In particular, the index statistic retains positional information discarded by PPV. Its larger feature representation therefore reflects additional pooling operations and an additional input representation, rather than simply more kernels.

\paragraph{Hydra.}
Hydra groups kernels into competing sets and counts which kernels achieve the strongest responses at each position~\cite{dempster2023hydra}. Hard counts record winning occurrences; soft counts also retain response magnitude. Features within a group are coupled by this competition, linking convolutional and dictionary-based representations. Hydra therefore illustrates how similar convolutional ingredients can generate different information through the pooling rule. Its competition mechanism is distinct from the independently pooled PPV map analysed here.

These methods share prescribed convolutional operations and a simple readout, while varying the parameter construction, input representation, and pooling. Our objective is to explain properties of the PPV map shared by the first three, without asserting that all guarantees extend to every family member.

\subsection{Invariance, stability, and random features}
\label{sec:int-related}
Invariance and stability are central to signal representations. Wavelet and scattering methods combine structured convolutions with nonlinearities to obtain translation invariance and stability to deformations~\cite{mallat2012wavelet,mallat2012group,bruna2013invariant}. Their architecture motivates related questions for PPV, but their guarantees do not directly transfer to randomly prescribed kernels followed by threshold pooling.

Random feature methods approximate a population representation using finitely many sampled features~\cite{rahimi2007random}. Random-weight neural networks and extreme learning machines provide related examples in which a useful representation is obtained without learning every parameter~\cite{huang2006universal,cao2018review}. Random projection theory offers another geometric viewpoint: appropriately sampled linear maps can preserve Euclidean distances on a finite dataset~\cite{bingham2001random,blum2005random}.

The PPV transform is nonlinear and need not preserve the original Euclidean distances. Its parameter distribution instead defines its own pairwise geometry. This distinction separates two questions: how faithfully a finite bank reproduces that population geometry, and how changing the distribution changes the geometry itself. The concentration analysis addresses the first; the kernel-length intervention addresses the second.

\section{The PPV Representation}
\label{sec:algorithm}
For kernel $k\in\{1,\ldots,K\}$, write $l_k$ for its length, $\mathbf w^{(k)}$ for its weights, $d^{(k)}$ for its dilation, $b^{(k)}$ for its bias, and $p_k$ for its symmetric zero-padding width. Let $\bar{\mathbf x}$ denote the zero extension of the input series. For finite boundaries the response is
\begin{equation}
 \mathbf u^{(k)}(\mathbf x)[n]
 =b^{(k)}+\sum_{j=1}^{l_k}w_j^{(k)}\bar{\mathbf x}[n-p_k+(j-1)d^{(k)}],
 \quad n=1,\ldots,N'_k,
 \label{eq:conv-output}
\end{equation}
where $N'_k=N+2p_k-(l_k-1)d^{(k)}\geq1$. Valid convolution has $p_k=0$. When $(l_k-1)d^{(k)}$ is even, choosing $p_k=(l_k-1)d^{(k)}/2$ preserves length. All experimental kernel lengths are odd, so this condition holds.

The PPV feature is the fraction of positive responses,
\begin{equation}
 \phi_k(\mathbf x)=\frac{1}{N'_k}\sum_{n=1}^{N'_k}
 \mathds1\!\left(\mathbf u^{(k)}(\mathbf x)[n]>0\right)\in[0,1].
 \label{eq:ppv}
\end{equation}
The feature map collects one coordinate per kernel:
\begin{equation}
 \Phi(\mathbf x)=[\phi_1(\mathbf x),\ldots,\phi_K(\mathbf x)]^\top\in\mathbb R^K.
 \label{eq:ppv-map}
\end{equation}
The bias controls the threshold: changing it probes a different part of the same convolutional response distribution. PPV retains how often a response exceeds that threshold but discards the positions of those occurrences.

These definitions separate the map from the procedure used to choose its parameters. The invariance result concerns any prescribed bank. The concentration results additionally assume independent draws from a common distribution. The geometric experiments use a standardised version of the raw map, defined in Section~\ref{subsec:experimental_protocol}.

\section{Translation Invariance and Stability}
\label{sec:translation_invariance}
PPV discards response locations, suggesting invariance to translations. The exact statement depends on boundary conditions: a circular translation reorders all responses under periodic boundaries, whereas finite translations interact with the observation window.
\subsection{Exact invariance under periodic boundaries}
\label{subsec:exact_translation}

Throughout this section $\mathbf{x}\in\mathbb{R}^{N}$ is extended $N$-periodically,
so that $\mathbf{x}[n]$ is defined for every $n\in\mathbb{Z}$ by
$\mathbf{x}[n]=\mathbf{x}[((n-1)\bmod N)+1]$, and indices need not be reduced
explicitly. Let $\tau_c\mathbf{x}$ denote the circular translation of
$\mathbf{x}$ by $c\in\mathbb{Z}$ samples,
\begin{equation}
    (\tau_c\mathbf{x})[n]
    =
    \mathbf{x}[n-c].
    \label{eq:circular_translation}
\end{equation}

Consider a prescribed kernel with weights
$\mathbf{w}^{(k)}\in\mathbb{R}^{l_k}$, dilation $d^{(k)}$ and bias $b^{(k)}$.
Under periodic boundary conditions its convolutional response, written
$\mathbf{u}^{(k)}(\mathbf{x})$ to make the dependence on the input explicit, is
\begin{equation}
    \mathbf{u}^{(k)}(\mathbf{x})[n]
    =
    \sum_{j=1}^{l_k}
    w^{(k)}_j\,
    \mathbf{x}[n+(j-1)d^{(k)}]
    +b^{(k)},
    \qquad n=1,\ldots,N .
    \label{eq:periodic_phi}
\end{equation}
The convolution preserves the input length, so $N'_k=N$.

\begin{proposition}[Exact translation invariance]
\label{prop:translation_invariance}
For every prescribed kernel $k$ and every circular translation
$c\in\mathbb{Z}$,
\begin{equation}
    \mathrm{PPV}\!\left(\mathbf{u}^{(k)}(\tau_c\mathbf{x})\right)
    =
    \mathrm{PPV}\!\left(\mathbf{u}^{(k)}(\mathbf{x})\right).
    \label{eq:ppv_invariance}
\end{equation}
\end{proposition}

\begin{proof}
Convolution is translation equivariant:
\begin{align}
    \mathbf{u}^{(k)}(\tau_c\mathbf{x})[n]
    &=
    \sum_{j=1}^{l_k}
    w^{(k)}_j\,
    \mathbf{x}[n+(j-1)d^{(k)}-c]
    +b^{(k)}
    \nonumber\\
    &=
    \mathbf{u}^{(k)}(\mathbf{x})[n-c].
    \label{eq:translation_equivariance}
\end{align}
Thus, translating the input circularly translates the convolutional response
without changing the set of response values. Applying PPV gives
\begin{equation}
    \mathrm{PPV}\!\left(\mathbf{u}^{(k)}(\tau_c\mathbf{x})\right)
    =
    \frac{1}{N}
    \sum_{n=1}^{N}
    \mathds{1}\!\left(
        \mathbf{u}^{(k)}(\mathbf{x})[n-c]>0
    \right).
\end{equation}
The mapping $n\mapsto n-c$ is a permutation of $\{1,\ldots,N\}$ modulo $N$. The
responses are therefore only reordered, and PPV is unchanged. Hence,
\begin{equation}
    \mathrm{PPV}\!\left(\mathbf{u}^{(k)}(\tau_c\mathbf{x})\right)
    =
    \mathrm{PPV}\!\left(\mathbf{u}^{(k)}(\mathbf{x})\right).
\end{equation}
\end{proof}

Since the result holds independently for every kernel, the whole representation
\eqref{eq:ppv-map} is invariant as well: $\Phi(\tau_c\mathbf{x})=\Phi(\mathbf{x})$.
The mechanism is simple. Convolution is translation equivariant, PPV is
insensitive to the order of the responses it summarises, and the composition of
the two is therefore translation invariant.

\subsection{Finite boundaries and general perturbations}
For finite signals, let $(\tau_c\mathbf x)[n]=\bar{\mathbf x}[n-c]$, so that samples leaving the observation interval are discarded and newly exposed positions are filled with zeros. With $R_k=(l_k-1)d^{(k)}<N$, the unpadded feature satisfies
\begin{equation}
 |\phi_k(\tau_c\mathbf x)-\phi_k(\mathbf x)|
 \leq\min\{1,|c|/(N-R_k)\}.
\end{equation}
For length-preserving zero padding, a bound is
\begin{equation}
 |\phi_k(\tau_c\mathbf x)-\phi_k(\mathbf x)|
 \leq\min\{1,(l_k+1)|c|/N\}.
\end{equation}
The proofs in Appendix~\ref{app:translation_stability} count unmatched or boundary-affected responses. These bounds clarify why exact periodic invariance does not imply exact invariance of the experimental implementation. They are informative when the affected responses form a small fraction of the pooling window.

Translation invariance is useful when a pattern's presence matters more than its location, but can discard discriminative latency information. MultiROCKET's mean-index pooling statistic explicitly complements PPV in this respect~\cite{tan2022multirocket}.

Hard thresholding also means that PPV need not be Lipschitz with respect to arbitrary input perturbations. Appendix~\ref{app:smooth_ppv_stability} instead analyses a sigmoid surrogate with sharpness $s>0$. Its per-feature sensitivity is bounded by $sG_k/(4\sqrt{N'_k})$, where $G_k$ is the maximum spectral gain of the kernel. This separates threshold sharpness, convolutional amplification, and the number of pooled responses. For centred unit-energy Gaussian kernels, the expected squared spectral gain is $O(\log l)$; the corresponding representation bound is derived in the appendix. The result concerns the smooth surrogate at fixed $s$ and does not establish a uniform Lipschitz guarantee for hard PPV.

\section{The Role of Random Parameters}
\label{sec:random_parameters}
For a fixed boundary convention and input length, the distribution of convolutional parameters induces a population geometry. A finite bank supplies a sampled approximation to it. This formulation distinguishes changing the number of kernels from changing their distribution.
\subsection{Population and empirical pairwise geometry}
\label{subsec:concentration}

Let $P$ denote the joint distribution from which ROCKET samples its kernel
parameters, including kernel length, weights, bias, dilation, and padding. Let
$\theta\sim P$ denote one such draw, and write
\begin{equation}
    \phi_\theta(\mathbf{x})
    \in[0,1]
\end{equation}
for its PPV feature.

For two time series $\mathbf{x}$ and $\mathbf{x}'$, define the population
separation
\begin{equation}
    \rho_P(\mathbf{x},\mathbf{x}')
    =
    \mathbb{E}_{\theta\sim P}
    \left[
        \left(
            \phi_\theta(\mathbf{x})
            -
            \phi_\theta(\mathbf{x}')
        \right)^2
    \right].
    \label{eq:population_distance}
\end{equation}
This quantity describes the pairwise geometry induced by the parameter
distribution $P$.

Given $K$ independent draws $\theta_1,\ldots,\theta_K\sim P$, define
\begin{align}
    \rho_K(\mathbf{x},\mathbf{x}')
    &=
    \frac{1}{K}
    \sum_{k=1}^{K}
    \left(
        \phi_{\theta_k}(\mathbf{x})
        -
        \phi_{\theta_k}(\mathbf{x}')
    \right)^2
    \nonumber\\
    &=
    \frac{1}{K}
    \left\|
        \Phi(\mathbf{x})
        -
        \Phi(\mathbf{x}')
    \right\|_2^2 .
    \label{eq:empirical_distance}
\end{align}

\begin{theorem}[Pointwise concentration of pairwise distance]
\label{thm:pointwise_concentration}
For every fixed pair $\mathbf{x},\mathbf{x}'$ and every $\varepsilon>0$,
\begin{equation}
    \mathbb{P}
    \left(
        |\rho_K(\mathbf{x},\mathbf{x}')
        -\rho_P(\mathbf{x},\mathbf{x}')|
        \geq\varepsilon
    \right)
    \leq
    2e^{-2K\varepsilon^2}.
\end{equation}
\end{theorem}

The result follows directly from Hoeffding's inequality because each squared
PPV difference lies in $[0,1]$.

\begin{corollary}[Uniform concentration on a finite dataset]
\label{cor:uniform_band}
Let $\mathbf{x}^{(1)},\ldots,\mathbf{x}^{(M)}$ be a dataset. Then, for any
$\delta\in(0,1)$, with probability at least $1-\delta$,
\begin{equation}
    \max_{1\leq i<i'\leq M}
    \left|
        \rho_K(\mathbf{x}^{(i)},\mathbf{x}^{(i')})
        -
        \rho_P(\mathbf{x}^{(i)},\mathbf{x}^{(i')})
    \right|
    \leq
    \sqrt{
        \frac{
            \log\!\left(
                2\binom{M}{2}/\delta
            \right)
        }{
            2K
        }
    }.
\end{equation}
\end{corollary}

The proofs are given in Appendix~\ref{app:concentration}. These results separate the roles of the parameter distribution and the number of sampled kernels. The distribution $P$ determines the population geometry, whereas $K$ determines how accurately a particular realisation of the transform approximates it.

The concentration results show that, as the number of kernels increases, different random realisations of the kernel bank induce increasingly similar pairwise geometries. This provides a form of reproducibility despite the randomness of the transform: for sufficiently large $K$, the representation depends less on the particular kernels that happened to be sampled. In turn, this suggests that downstream results should become more consistent across different realisations of ROCKET, provided that small variations in the induced geometry do not substantially affect the classifier. Such reproducibility is a highly desirable property for an algorithm based on random features.

This is a concentration statement for squared distances in the raw PPV feature space, normalised by $K$. It is not a Johnson--Lindenstrauss guarantee for the original input distances, a classification guarantee, or a direct bound for geometry after data-dependent standardisation. Those distinctions matter when relating the theorem to the empirical measures below.

\section{Experimental Protocol}
\label{subsec:experimental_protocol}
\subsection{Implementation and datasets}
We use a PPV-only adaptation of the reference Python implementation of ROCKET and the standard 112-dataset classification collection from the UCR archive~\cite{dau2019ucr}. After mean-centring, each kernel is normalised to unit Euclidean norm. This fixes its total squared weight, or energy, across kernel lengths, controlling a source of variation in convolution-response scale. Biases remain uniform on $[-1,1]$, and dilation and padding follow the sampling rules described in Section~\ref{sec:family}.

Each dataset and configuration uses one independently sampled kernel bank. We retain the archive's standard train/test partitions. The same fitted bank transforms both splits. For each PPV coordinate, we estimate its mean and standard deviation on the training set, using a scale of one for zero training variance, and apply the same transformation to the test set. Let $\widetilde\Phi$ denote this standardised map. Both geometric quantities are calculated on the standardised training features; the classifier uses the standardised features in both splits.

Classification uses a ridge classifier with regularisation selected from ten logarithmically spaced values between $10^{-3}$ and $10^3$, using training-data leave-one-out cross-validation with accuracy as the selection score. We record both training and test accuracy. The implementation and analysis code, configurations, metadata, and result files are archived in version~1.0.0~\cite{marco_blanco2026kernelsoftware}; the development repository is \url{https://github.com/jorgemarcoes/prescribed-kernel-interpretability}.

The concentration bounds concern the raw map $\Phi$. For illustration, with $K=10{,}000$, $M=1{,}000$, and $\delta=0.05$, the uniform bound is approximately $0.029$. Its size relative to the population separations depends on their scale, so it does not by itself establish that one realisation is sufficiently accurate for every dataset. The empirical analysis is descriptive and does not estimate variation over repeated kernel banks.

\subsection{Geometric quantities and aggregation}
For the covariance eigenvalues $\mu_j$ of the standardised training feature matrix, effective dimensionality is the participation ratio
\begin{equation}
 \mathrm{ED}=\frac{(\sum_j\mu_j)^2}{\sum_j\mu_j^2}.
 \label{eq:participation_ratio}
\end{equation}
It measures how broadly the representation's variance is distributed over directions, rather than simply counting its ambient coordinates. Numerical zero eigenvalues are discarded; an entirely zero-variance representation is assigned ED zero.

Distance correlation~\cite{szekely2007measuring} compares the input series with their standardised representations. Let
\begin{equation}
 a_{ij}=\|\mathbf x^{(i)}-\mathbf x^{(j)}\|_2,\qquad
 b_{ij}=\|\widetilde\Phi(\mathbf x^{(i)})-\widetilde\Phi(\mathbf x^{(j)})\|_2,
\end{equation}
and double-centre these matrices, for example
$A_{ij}=a_{ij}-\bar a_{i\cdot}-\bar a_{\cdot j}+\bar a_{\cdot\cdot}$, with $B$ defined analogously. The sample distance correlation is
\begin{equation}
 \mathrm{dCor}(X,\widetilde\Phi(X))=
 \frac{(M^{-2}\sum_{i,j}A_{ij}B_{ij})^{1/2}}
 {[(M^{-2}\sum_{i,j}A_{ij}^2)(M^{-2}\sum_{i,j}B_{ij}^2)]^{1/4}}.
 \label{eq:distance_correlation}
\end{equation}
The implementation returns zero if either distance variance is zero. All training observations are used, without subsampling.

We report absolute metrics and within-dataset relative metrics. For dataset index $r$ and configuration $c$,
\begin{equation}
 \mathrm{RelED}_{r,c}=\frac{\mathrm{ED}_{r,c}}{\max_{c'}\mathrm{ED}_{r,c'}},
\end{equation}
with relative dCor defined analogously. The denominator is the maximum observed within the corresponding experiment, not a theoretical maximum. Curves show means across datasets with pointwise 95\% bootstrap confidence intervals, using 1,000 dataset resamples. These intervals concern the across-dataset mean and do not quantify variability over independent kernel-bank repetitions.

\subsection{Experimental interventions}
The kernel-count experiment varies $K\in\{100,500,1000,2000,5000,10000\}$, with kernel lengths sampled uniformly from $\{7,9,11\}$. The parameter distribution remains fixed, and all 112 datasets contribute at every count.

The kernel-length experiment fixes $K=10{,}000$ and sets every kernel in a run to the same odd length $l\in\{3,5,\ldots,27\}$. This changes one component of the parameter distribution while retaining the same ambient feature dimension. Every kernel-length analysis uses the same 109 datasets with $N\geq27$, excluding SmoothSubspace, Chinatown, and ItalyPowerDemand at every length. The restriction prevents changes in dataset composition across configurations; it applies to both geometric metrics and classification summaries.

For the classification comparison, we rank the 13 lengths separately by training and test accuracy within each dataset. Rank one denotes the highest accuracy and tied accuracies receive average ranks. We then average ranks and absolute accuracies over the same 109 datasets. Kernel length is an experimental intervention here; the reported test results describe all configurations rather than a separately validated procedure for choosing a length on a new dataset.

\section{Results}
\subsection{Number of kernels}
\label{subsec:ed_vs_kernels}
\begin{figure}[H]
\centering
\includegraphics[width=\linewidth]{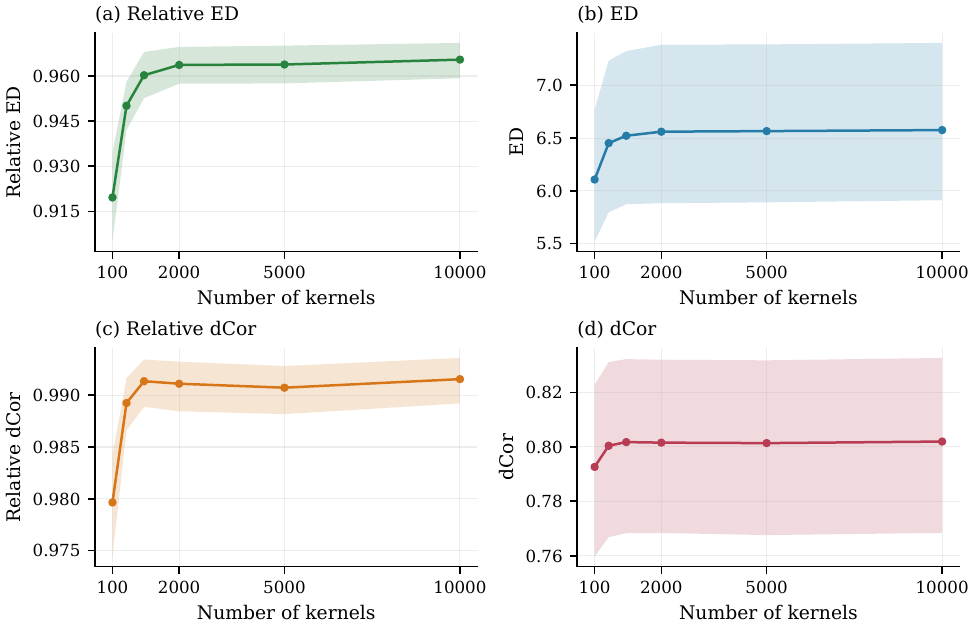}
\caption{Geometry versus kernel count, averaged over all 112 UCR datasets. Panels show relative and absolute ED and distance correlation for standardised training PPV features. Relative values are normalised by each dataset's maximum within this experiment. Shading denotes pointwise 95\% bootstrap confidence intervals across datasets.}
\label{fig:kernel-count}
\end{figure}

Figure~\ref{fig:kernel-count} shows that mean ED increases with the number of kernels, with the largest gains at small counts and progressively smaller changes thereafter. The absolute and relative summaries give complementary views: the former retains the scale of each dataset, whereas the latter gives equal emphasis to proportional changes relative to the dataset's observed maximum.

Distance correlation changes comparatively little after its initial rise. Thus, a relatively small bank captures much of the observed dependence between input and transformed pairwise geometries, while further kernels continue to increase the effective dimensionality of the finite representation. Additional coordinates can enrich the representation even after its broad relationship with the input geometry changes only modestly.

This pattern is compatible with the diminishing classification returns reported in the original ROCKET sensitivity analysis~\cite{dempster2020rocket}. Here, however, the measured quantities concern representation geometry. The concentration theorem describes approximation of the population PPV geometry; it does not require ED or dCor to increase monotonically for every finite dataset.

\subsection{Kernel length and representation geometry}
\label{sec:kernel_length}
\begin{figure}[H]
\centering
\includegraphics[width=\linewidth]{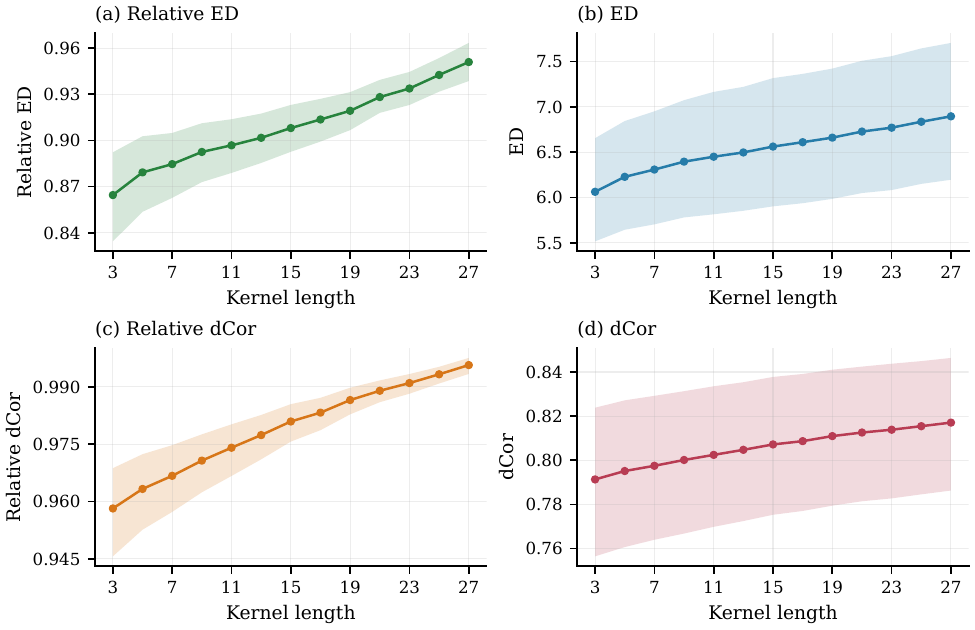}
\caption{Geometry versus kernel length with $10{,}000$ kernels, averaged over the same 109 UCR datasets with $N\geq27$ at every length. Panels show relative and absolute ED and distance correlation for standardised training PPV features. Relative values use each dataset's maximum within this experiment. Shading denotes pointwise 95\% bootstrap confidence intervals across datasets.}
\label{fig:kernel-length}
\end{figure}

Figure~\ref{fig:kernel-length} examines a different intervention: the feature count is fixed while the convolutional parameter distribution changes. Mean relative ED and mean relative dCor increase across the tested lengths. The absolute summaries show the same overall trend on the fixed 109-dataset cohort.

The ED increase indicates that longer kernels distribute feature variance over more effective directions, despite keeping the ambient dimension constant. Longer kernels contain more coefficients and can aggregate information across broader temporal contexts, particularly with dilation. This can change the diversity and redundancy of convolutional responses. At the same time, increasing dCor indicates stronger dependence between the input and transformed geometries. The two quantities therefore describe complementary effects rather than interchangeable measures of representation quality.

Both geometric summaries continue to increase beyond the standard ROCKET kernel-length range. Intermediate lengths consequently do not maximise these geometric quantities in the tested range. The next comparison asks whether the richer representations also improve classification.

\subsection{Kernel length and classification performance}
\label{subsec:length_accuracy}
\begin{figure}[H]
\centering
\includegraphics[width=0.82\linewidth]{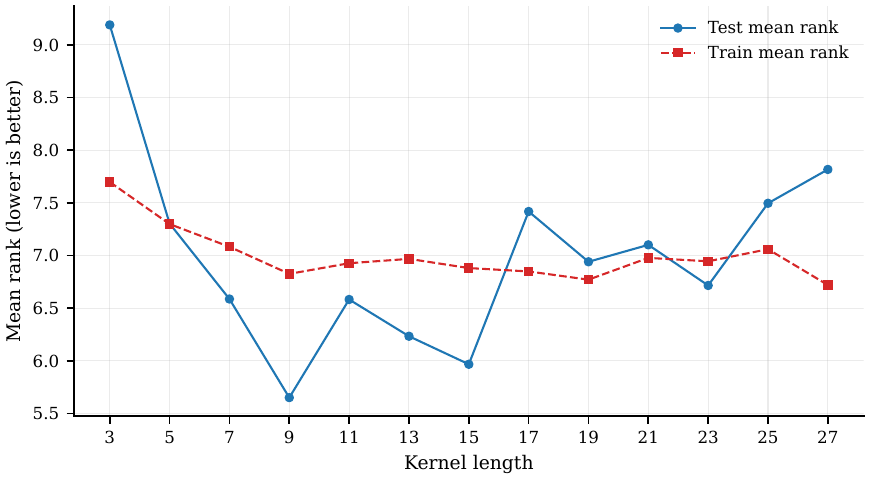}
\caption{Training and test mean ranks across the same 109 UCR datasets. The 13 lengths are ranked separately by training and test accuracy within each dataset, with average ranks for ties. Lower ranks indicate better relative performance.}
\label{fig:train_vs_test_mean_rank}
\end{figure}

Figure~\ref{fig:train_vs_test_mean_rank} compares mean training and test ranks, and Table~\ref{tab:kernel_length_train_test} supplies the corresponding absolute accuracies. Test mean rank improves from the shortest kernels to an intermediate regime, attaining its lowest observed value at length 9. At longer lengths, test ranks generally worsen, although the pattern is not monotonic. Training ranks vary less and remain comparatively favourable for several longer configurations.

\begin{table}[H]
\centering
\caption{Mean training and test ranks and mean absolute accuracies for the
13 kernel lengths across the same 109 UCR datasets with $N\geq27$.
Lower ranks and higher accuracies indicate better performance. Tied accuracies
receive average ranks within each dataset.}
\label{tab:kernel_length_train_test}
\begin{tabular}{@{}ccccc@{}}
\toprule
Kernel length & Train rank & Test rank & Train accuracy & Test accuracy \\
\midrule
3 & 7.70 & 9.19 & 0.9958 & 0.8298 \\
5 & 7.30 & 7.30 & 0.9966 & 0.8424 \\
7 & 7.08 & 6.59 & 0.9977 & 0.8445 \\
9 & 6.83 & 5.65 & 0.9982 & 0.8481 \\
11 & 6.93 & 6.58 & 0.9982 & 0.8456 \\
13 & 6.97 & 6.23 & 0.9971 & 0.8448 \\
15 & 6.88 & 5.97 & 0.9983 & 0.8460 \\
17 & 6.85 & 7.42 & 0.9978 & 0.8434 \\
19 & 6.77 & 6.94 & 0.9984 & 0.8433 \\
21 & 6.98 & 7.10 & 0.9978 & 0.8439 \\
23 & 6.94 & 6.72 & 0.9982 & 0.8452 \\
25 & 7.06 & 7.50 & 0.9977 & 0.8438 \\
27 & 6.72 & 7.82 & 0.9981 & 0.8430 \\
\bottomrule
\end{tabular}
\end{table}
Training accuracy is close to one and frequently tied, so small absolute differences can change ranks while ties can obscure changes in training fit. The table provides the scale needed to interpret that comparison. The additional representational richness at longer lengths does not translate into better relative test performance in this experiment.

ROCKET's default lengths $\{7,9,11\}$ and MiniROCKET's length 9 lie within the favourable intermediate region~\cite{dempster2020rocket,dempster2021minirocket}. This provides a retrospective validation of those design choices: lengths selected empirically in the original methods fall within the region of strongest relative test performance in the present analysis. It does not imply that one length is optimal for every dataset or that the modified transform reproduces the original algorithms' accuracies exactly.
\FloatBarrier

\section{Discussion}
\subsection{A geometric interpretation}
The theoretical and empirical results separate several contributions to the representation. Convolution detects patterns at multiple positions and scales; the bias thresholds each response; and PPV summarises the frequency with which that threshold is exceeded. The resulting nonlinearity can distribute observations over directions that a linear classifier could not access directly in the input space. Periodic translation invariance follows from discarding response locations, while the finite-boundary bounds identify when that invariance is only approximate.

The population viewpoint explains a different part of the construction. Once a distribution of convolutional parameters is fixed, its expected squared PPV separation defines a pairwise geometry. Increasing the number of independent kernels improves concentration around this geometry. Randomness supplies one way of sampling a useful parameter space; it is not itself a proof that the geometry is useful for classification. MiniROCKET's prescribed construction is consistent with the possibility that structured parameter sets can provide useful coverage without relying on independent random weights.

Effective dimensionality helps distinguish a large coordinate count from an expansion over effective directions. The observed ED increases are consistent with the consequence of Cover's theorem~\cite{cover1965geometrical}: a nonlinear embedding into a sufficiently high-dimensional, non-densely populated space can make a complex classification problem more amenable to linear separation. Distance correlation adds evidence that the transformed representation remains statistically related to the original signals. Together, the measures support an interpretation of a richer but structured geometry; neither measure alone establishes class separability or distance preservation.

\subsection{Generalisation and limits of the evidence}
Longer kernels generally increase the geometric measures while intermediate lengths yield stronger test ranks. This is compatible with the familiar bias--variance intuition that additional effective directions can increase the classifier's flexibility without improving generalisation. The divergence of training and test ranks at several longer lengths supports discussing possible overfitting, with the qualification that training accuracies are near saturation and often tied. The comparison does not isolate the mechanism responsible for the test behaviour.

Several boundaries of the analysis are explicit. The exact invariance theorem assumes periodic boundaries, whereas the implementation mixes valid and zero-padded convolution. The perturbation bound concerns smooth PPV and does not extend uniformly to the hard threshold. Concentration requires independent draws from a common parameter distribution and concerns the raw PPV map, while the empirical metrics use data-dependent standardisation. It therefore does not establish the same guarantee for the measured standardised geometry, or directly for prescribed and coupled constructions such as MiniROCKET and Hydra.

The experiments use one random bank per configuration. Across-dataset confidence intervals summarise the means for the benchmark collection; they do not establish a per-dataset monotonic trend or quantify repeated-bank uncertainty. Unit-energy normalisation also changes the reference ROCKET distribution, so the experiments should be read as an intervention on this stated PPV-only construction. Holding the cohort fixed prevents composition changes from explaining the kernel-length curves, but does not make geometric richness a sufficient criterion for selecting a classifier representation.

\section{Conclusions}
Prescribed-parameter convolutional transforms can be interpreted through properties of the representation they produce, independently of the simple linear readout fitted afterwards. PPV converts translation-equivariant responses into exactly invariant features under periodic boundaries, with finite-boundary departures controlled by the affected response locations. A distribution of random convolutional parameters defines a population pairwise geometry, and a finite independent kernel bank approximates it with concentration improving as the bank grows.

The empirical interventions distinguish approximation from a change in representation. Increasing kernel count produces diminishing geometric gains. Changing kernel length changes the geometry itself: longer kernels generally increase ED and dCor, while intermediate lengths achieve stronger test performance. This supports a mechanistic interpretation in which nonlinear pooling produces additional effective directions that can favour linear separation, while greater representational richness can offer progressively less benefit for generalisation.

Open directions include stability of hard PPV under noise and deformations, repeated-bank variability, and the relationship between parameter distributions and class geometry.

\FloatBarrier
\bibliographystyle{unsrtnat}
\bibliography{references}

\begin{thebibliography}{18}
\providecommand{\natexlab}[1]{#1}
\providecommand{\url}[1]{\texttt{#1}}
\expandafter\ifx\csname urlstyle\endcsname\relax
  \providecommand{\doi}[1]{doi: #1}\else
  \providecommand{\doi}{doi: \begingroup \urlstyle{rm}\Url}\fi

\bibitem[Dempster et~al.(2020)Dempster, Petitjean, and
  Webb]{dempster2020rocket}
Angus Dempster, Fran{\c{c}}ois Petitjean, and Geoffrey~I Webb.
\newblock {ROCKET}: exceptionally fast and accurate time series classification
  using random convolutional kernels.
\newblock \emph{Data Mining and Knowledge Discovery}, 34\penalty0 (5):\penalty0
  1454--1495, 2020.

\bibitem[Dempster et~al.(2021)Dempster, Schmidt, and
  Webb]{dempster2021minirocket}
Angus Dempster, Daniel~F Schmidt, and Geoffrey~I Webb.
\newblock {MiniROCKET}: A very fast (almost) deterministic transform for time
  series classification.
\newblock In \emph{Proceedings of the 27th ACM SIGKDD conference on knowledge
  discovery \& data mining}, pages 248--257, 2021.

\bibitem[Tan et~al.(2022)Tan, Dempster, Bergmeir, and Webb]{tan2022multirocket}
Chang~Wei Tan, Angus Dempster, Christoph Bergmeir, and Geoffrey~I Webb.
\newblock {MultiROCKET}: multiple pooling operators and transformations for
  fast and effective time series classification.
\newblock \emph{Data Mining and Knowledge Discovery}, 36\penalty0 (5):\penalty0
  1623--1646, 2022.

\bibitem[Dempster et~al.(2023)Dempster, Schmidt, and Webb]{dempster2023hydra}
Angus Dempster, Daniel~F Schmidt, and Geoffrey~I Webb.
\newblock {Hydra}: competing convolutional kernels for fast and accurate time
  series classification.
\newblock \emph{Data Mining and Knowledge Discovery}, 37\penalty0 (5):\penalty0
  1779--1805, 2023.

\bibitem[Cover(1965)]{cover1965geometrical}
Thomas~M. Cover.
\newblock Geometrical and statistical properties of systems of linear
  inequalities with applications in pattern recognition.
\newblock \emph{IEEE Transactions on Electronic Computers}, EC-14\penalty0
  (3):\penalty0 326--334, June 1965.
\newblock \doi{10.1109/PGEC.1965.264137}.

\bibitem[Hoerl and Kennard(1970)]{hoerl1970ridge}
Arthur~E Hoerl and Robert~W Kennard.
\newblock Ridge regression: Biased estimation for nonorthogonal problems.
\newblock \emph{Technometrics}, 12\penalty0 (1):\penalty0 55--67, 1970.

\bibitem[Hastie et~al.(2009)Hastie, Tibshirani, and
  Friedman]{hastie2009elements}
Trevor Hastie, Robert Tibshirani, and Jerome Friedman.
\newblock \emph{The Elements of Statistical Learning: Data Mining, Inference,
  and Prediction}.
\newblock Springer, 2 edition, 2009.
\newblock \doi{10.1007/978-0-387-84858-7}.

\bibitem[Mallat(2012{\natexlab{a}})]{mallat2012wavelet}
St\'{e}phane Mallat.
\newblock \emph{A Wavelet Tour of Signal Processing: The Sparse Way}.
\newblock Academic Press, 2012{\natexlab{a}}.
\newblock ISBN 978-0-12-394424-4.

\bibitem[Mallat(2012{\natexlab{b}})]{mallat2012group}
St{\'e}phane Mallat.
\newblock Group invariant scattering.
\newblock \emph{Communications on Pure and Applied Mathematics}, 65\penalty0
  (10):\penalty0 1331--1398, 2012{\natexlab{b}}.

\bibitem[Bruna and Mallat(2013)]{bruna2013invariant}
Joan Bruna and St\'{e}phane Mallat.
\newblock Invariant scattering convolution networks.
\newblock \emph{IEEE Transactions on Pattern Analysis and Machine
  Intelligence}, 35\penalty0 (8):\penalty0 1872--1886, 2013.
\newblock \doi{10.1109/TPAMI.2012.57}.
\newblock URL \url{https://ieeexplore.ieee.org/document/6298094}.

\bibitem[Rahimi and Recht(2007)]{rahimi2007random}
Ali Rahimi and Benjamin Recht.
\newblock Random features for large-scale kernel machines.
\newblock \emph{Advances in neural information processing systems}, 20, 2007.

\bibitem[Huang et~al.(2006)Huang, Chen, and Siew]{huang2006universal}
Guang-Bin Huang, Lei Chen, and Chee-Kheong Siew.
\newblock Universal approximation using incremental constructive feedforward
  networks with random hidden nodes.
\newblock \emph{IEEE transactions on neural networks}, 17\penalty0
  (4):\penalty0 879--892, 2006.

\bibitem[Cao et~al.(2018)Cao, Wang, Ming, and Gao]{cao2018review}
Weipeng Cao, Xizhao Wang, Zhong Ming, and Jinzhu Gao.
\newblock A review on neural networks with random weights.
\newblock \emph{Neurocomputing}, 275:\penalty0 278--287, 2018.

\bibitem[Bingham and Mannila(2001)]{bingham2001random}
Ella Bingham and Heikki Mannila.
\newblock Random projection in dimensionality reduction: applications to image
  and text data.
\newblock In \emph{Proceedings of the seventh ACM SIGKDD international
  conference on Knowledge discovery and data mining}, pages 245--250, 2001.

\bibitem[Blum(2005)]{blum2005random}
Avrim Blum.
\newblock Random projection, margins, kernels, and feature-selection.
\newblock In \emph{International Statistical and Optimization Perspectives
  Workshop" Subspace, Latent Structure and Feature Selection"}, pages 52--68.
  Springer, 2005.

\bibitem[Dau et~al.(2019)Dau, Bagnall, Kamgar, Yeh, Zhu, Gharghabi,
  Ratanamahatana, and Keogh]{dau2019ucr}
Hoang~Anh Dau, Anthony Bagnall, Kaveh Kamgar, Chin-Chia~Michael Yeh, Yan Zhu,
  Shaghayegh Gharghabi, Chotirat~Ann Ratanamahatana, and Eamonn Keogh.
\newblock The ucr time series archive.
\newblock \emph{IEEE/CAA Journal of Automatica Sinica}, 6\penalty0
  (6):\penalty0 1293--1305, 2019.

\bibitem[Marco~Blanco(2026)]{marco_blanco2026kernelsoftware}
Jorge Marco~Blanco.
\newblock A mechanistic interpretation of prescribed-parameter convolutional
  kernels: Companion code.
\newblock Computer software, version v1.0.0. Zenodo, October 2026.
\newblock URL \url{https://doi.org/10.5281/zenodo.23672114}.

\bibitem[Sz{\'e}kely et~al.(2007)Sz{\'e}kely, Rizzo, and
  Bakirov]{szekely2007measuring}
G{\'a}bor~J Sz{\'e}kely, Maria~L Rizzo, and Nail~K Bakirov.
\newblock Measuring and testing dependence by correlation of distances.
\newblock \emph{The Annals of Statistics}, 35\penalty0 (6):\penalty0
  2769--2794, 2007.
\newblock \doi{10.1214/009053607000000505}.

\end{thebibliography}
\clearpage
\appendix
\numberwithin{equation}{section}
\section{Translation stability under finite boundary conditions}
\label{app:translation_stability}

The exact result in Section~\ref{sec:translation_invariance} assumes periodic
boundary conditions. In practice, a finite time series is not periodic, and
translations may move samples outside the observed interval. Consequently, PPV
is not exactly translation invariant. Nevertheless, its deviation from
invariance can be controlled by the number of convolutional responses affected
by the boundaries.

Throughout we use the notation of Section~\ref{sec:algorithm}: the $k$-th
kernel has weights $\mathbf{w}^{(k)}\in\mathbb{R}^{l_k}$, dilation $d^{(k)}$
and bias $b^{(k)}$; $\mathbf{u}^{(k)}(\mathbf{x})$ denotes its convolutional
response to the input $\mathbf{x}\in\mathbb{R}^{N}$, and
$\phi_k(\mathbf{x})=\mathrm{PPV}(\mathbf{u}^{(k)}(\mathbf{x}))$ the
corresponding feature.

Let the zero extension of $\mathbf{x}$ be
\begin{equation}
    \bar{\mathbf{x}}[n]
    =
    \begin{cases}
        \mathbf{x}[n], & 1\leq n\leq N,\\
        0,             & \text{otherwise},
    \end{cases}
\end{equation}
and define a finite translation by
\begin{equation}
    (\tau_c\mathbf{x})[n]
    =
    \bar{\mathbf{x}}[n-c],
    \qquad 1\leq n\leq N.
    \label{eq:finite_translation}
\end{equation}
Samples shifted outside the observation interval are therefore discarded, while
newly exposed positions are filled with zeros.

For kernel $k$, let
\begin{equation}
    R_k=(l_k-1)d^{(k)}
    \label{eq:effective_span}
\end{equation}
denote its effective convolutional span, so that its receptive field spans
$R_k+1$ samples, as in Section~\ref{sec:conv1d}.

\subsection{Unpadded convolution}
\label{app:unpadded}

Suppose first that no padding is applied. The convolution contains
\begin{equation}
    N'_k=N-R_k
\end{equation}
valid output positions, and the corresponding PPV feature is
\begin{equation}
    \phi_k(\mathbf{x})
    =
    \frac{1}{N'_k}
    \sum_{n=1}^{N'_k}
    \mathds{1}\!\left(
    \mathbf{u}^{(k)}(\mathbf{x})[n]>0
    \right).
\end{equation}

\begin{proposition}[Translation stability, unpadded convolution]
\label{prop:unpadded_stability}
For $|c|<N$,
\begin{equation}
    \left|
    \phi_k(\tau_c\mathbf{x})-\phi_k(\mathbf{x})
    \right|
    \leq
    \min\left\{
        1,\,
        \frac{|c|}{N-R_k}
    \right\}.
    \label{eq:unpadded_bound}
\end{equation}
\end{proposition}

\begin{proof}
Away from the boundaries, convolution remains translation equivariant, so the
response obtained from the translated signal coincides with a translated
version of the original convolutional response. The two PPV sums therefore
agree on all indices common to their respective convolution windows.
Translating a window of length $N'_k$ by $|c|$ positions leaves at most $|c|$
unmatched terms. Since each indicator takes a value in $\{0,1\}$, the
difference between the two sums is at most $|c|$. Dividing by $N'_k=N-R_k$
yields \eqref{eq:unpadded_bound}.
\end{proof}

Thus, an unpadded PPV feature is close to translation invariant whenever
\begin{equation}
    |c|\ll N-R_k.
\end{equation}
Large dilations increase $R_k$ and reduce the number of valid convolution
positions, making the corresponding feature more sensitive to translations.

\subsection{Length-preserving zero padding}
\label{app:padded}

Suppose now that the convolution is zero padded so that
$\mathbf{u}^{(k)}(\mathbf{x})$ has length $N$, as in the notation of
Section~\ref{sec:translation_invariance}. A translated input may differ from an
infinite translation only through the samples discarded at the signal boundary.

Each discarded input sample can contribute to at most $l_k$ convolutional
responses, one through each coefficient of $\mathbf{w}^{(k)}$. A translation by
$|c|$ samples discards at most $|c|$ input samples. In addition, translating
the output window produces at most $|c|$ unmatched response locations.

\begin{proposition}[Translation stability, padded convolution]
\label{prop:padded_stability}
For a length-preserving zero-padded convolution,
\begin{equation}
    \left|
    \phi_k(\tau_c\mathbf{x})-\phi_k(\mathbf{x})
    \right|
    \leq
    \min\left\{
        1,\,
        \frac{(l_k+1)|c|}{N}
    \right\}.
    \label{eq:padded_bound}
\end{equation}
\end{proposition}

\begin{proof}
At most $l_k|c|$ convolutional responses can be altered by the samples
discarded at the boundary, while at most $|c|$ additional responses arise from
the displacement of the finite pooling window. Hence, at most $(l_k+1)|c|$
indicator terms can differ between the two PPV sums. Since each indicator
differs by at most one,
\begin{equation}
    \left|
    \sum_{n=1}^{N}
    \mathds{1}\!\left(
    \mathbf{u}^{(k)}(\tau_c\mathbf{x})[n]>0
    \right)
    -
    \sum_{n=1}^{N}
    \mathds{1}\!\left(
    \mathbf{u}^{(k)}(\mathbf{x})[n]>0
    \right)
    \right|
    \leq
    (l_k+1)|c|.
\end{equation}
Dividing by $N$ and using the trivial upper bound of one gives
\eqref{eq:padded_bound}.
\end{proof}

Propositions~\ref{prop:unpadded_stability} and~\ref{prop:padded_stability}
clarify the relation between the periodic model and the finite transform. PPV
pooling is exactly invariant under circular translations, whereas under finite
boundary conditions it is \emph{translation stable}: the discrepancy is
confined to responses affected by the boundaries. In particular, for
length-preserving padding, the deviation is small whenever
\begin{equation}
    l_k|c|\ll N.
\end{equation}

For the complete representation
$\Phi(\mathbf{x})\in\mathbb{R}^{K}$ of~\eqref{eq:ppv-map}, let
$\epsilon_k(c)$ denote the corresponding bound from
Proposition~\ref{prop:unpadded_stability} or
Proposition~\ref{prop:padded_stability}. Then
\begin{equation}
    \frac{1}{K}
    \left\|
        \Phi(\tau_c\mathbf{x})-\Phi(\mathbf{x})
    \right\|_1
    \leq
    \frac{1}{K}
    \sum_{k=1}^{K}\epsilon_k(c).
    \label{eq:feature_vector_stability}
\end{equation}
Therefore, the finite PPV representation should not be regarded as exactly
translation invariant; rather, it approaches the invariant periodic model when
boundary-affected responses form only a small fraction of the pooled
convolutional outputs.

\section{Smooth-PPV stability and unit-energy spectral scaling}
\label{app:smooth_ppv_stability}

Hard PPV is discontinuous because of its threshold. We therefore replace the
indicator by a sigmoid and study a smooth surrogate.

For a prescribed kernel $k$, let $p_k$ denote the left-padding width
($p_k=0$ for valid convolution). Let $\bar{\mathbf{x}}$ be the zero extension
of $\mathbf{x}$: it equals $\mathbf{x}[q]$ for $1\leq q\leq N$ and zero
otherwise. Write
\begin{equation}
    a_{k,n}(\mathbf{x})
    =
    \sum_{j=1}^{l_k}
    w^{(k)}_j\,
    \bar{\mathbf{x}}[n-p_k+(j-1)d^{(k)}]
    +b^{(k)},
    \qquad
    n=1,\ldots,N'_k,
\end{equation}
for the $n$-th convolutional response, that is, $a_{k,n}(\mathbf{x})
=\mathbf{u}^{(k)}[n]$, with the dependence on the input made explicit.

For $s>0$, let
\begin{equation}
    \sigma_s(u)
    =
    \frac{1}{1+e^{-su}},
\end{equation}
and define
\begin{equation}
    \phi_{s,k}(\mathbf{x})
    =
    \frac{1}{N'_k}
    \sum_{n=1}^{N'_k}
    \sigma_s\!\left(a_{k,n}(\mathbf{x})\right).
\end{equation}

The smooth representation is
\begin{equation}
    \Phi_s(\mathbf{x})
    =
    \left[
        \phi_{s,1}(\mathbf{x}),
        \ldots,
        \phi_{s,K}(\mathbf{x})
    \right]^\top .
\end{equation}

\subsection*{Main stability result}
\label{subsec:main_stability}

We express stability in terms of Lipschitz continuity. A map
$F:\mathbb{R}^{N}\to\mathbb{R}^{m}$ is $L$-Lipschitz if
\begin{equation}
\|F(\mathbf{y})-F(\mathbf{x})\|_2
\leq
L\|\mathbf{y}-\mathbf{x}\|_2
\end{equation}
for all inputs $\mathbf{x}$ and $\mathbf{y}$. The constant $L$ therefore
provides a global upper bound on how strongly perturbations of the input can be
amplified by the representation. Smaller Lipschitz constants correspond to
stronger stability. We consider valid convolution or length-preserving zero-padded convolution, with positive integer dilation and \(N'_k\geq1\).

\begin{proposition}[Stability of smooth PPV features]
\label{prop:smooth_stability}

For a prescribed kernel $k$, define its maximum spectral gain as
\begin{equation}
G_k
=
\sup_{\omega}
\left|
\sum_{j=1}^{l_k}
w_j^{(k)}
e^{-\mathrm{i}\omega(j-1)}
\right|.
\label{eq:spectral_gain}
\end{equation}
Then, for any two input signals $\mathbf{x}$ and $\mathbf{y}$:

\emph{(i) Per-feature stability.}
\begin{equation}
    |\phi_{s,k}(\mathbf{y})
      -\phi_{s,k}(\mathbf{x})|
    \le
    \frac{s\,G_k}{4\sqrt{N'_k}}
    \|\mathbf{y}-\mathbf{x}\|_2 .
    \label{eq:per_feature_stability}
\end{equation}

\emph{(ii) Representation stability.}
\begin{equation}
    \|\Phi_s(\mathbf{y})
      -\Phi_s(\mathbf{x})\|_2
    \le
    C_s\|\mathbf{y}-\mathbf{x}\|_2 ,
    \label{eq:lipschitz}
\end{equation}
where
\begin{equation}
    C_s
    =
    \frac{s}{4}
    \left(
        \sum_{k=1}^{K}
        \frac{G_k^2}{N'_k}
    \right)^{1/2}.
    \label{eq:stability_constant}
\end{equation}
\end{proposition}

\begin{proof}
    
\label{app:stability-proof}

The sigmoid derivative satisfies
\begin{equation}
    \sigma_s'(u)
    =
    s\,
    \sigma_s(u)
    \bigl(1-\sigma_s(u)\bigr),
\end{equation}
and therefore
\begin{equation}
    |\sigma_s'(u)|
    \le
    \frac{s}{4}.
    \label{eq:sigmoid_lipschitz}
\end{equation}

Hence, for kernel $k$,
\begin{equation}
    |\phi_{s,k}(\mathbf{y})-\phi_{s,k}(\mathbf{x})|
    \le
    \frac{s}{4N'_k}
    \sum_{n=1}^{N'_k}
    |\mathbf{v}^{(k)}[n]|,
    \label{eq:feature_first_bound}
\end{equation}
where $\mathbf{v}^{(k)}$ is the difference between the convolutional responses to $\mathbf{y}$ and $\mathbf{x}$. The bias cancels, so this is the linear convolution of $\mathbf{y}-\mathbf{x}$ with the kernel weights.

By Cauchy--Schwarz,
\begin{equation}
    \sum_{n=1}^{N'_k}|\mathbf{v}^{(k)}[n]|
    \le
    \sqrt{N'_k}\,\|\mathbf{v}^{(k)}\|_2,
\end{equation}
so
\begin{equation}
    |\phi_{s,k}(\mathbf{y})-\phi_{s,k}(\mathbf{x})|
    \le
    \frac{s}{4\sqrt{N'_k}}
    \|\mathbf{v}^{(k)}\|_2.
    \label{eq:averaging_bound}
\end{equation}

For discrete convolution, the $\ell_2$ amplification is bounded by the maximum
magnitude of the frequency response, that is, by the spectral gain $G_k$ of
\eqref{eq:spectral_gain},
\begin{equation}
    \|\mathbf{v}^{(k)}\|_2
    \le
    G_k\|\mathbf{y}-\mathbf{x}\|_2.
    \label{eq:spectral_bound}
\end{equation}
Dilation reparameterises the frequency response and does not change its
supremum.

Combining \eqref{eq:averaging_bound} and \eqref{eq:spectral_bound} gives
\begin{equation}
    |\phi_{s,k}(\mathbf{y})-\phi_{s,k}(\mathbf{x})|
    \le
    \frac{s\,G_k}{4\sqrt{N'_k}}
    \|\mathbf{y}-\mathbf{x}\|_2,
\end{equation}
which proves part~(i) of Proposition~\ref{prop:smooth_stability}.

For part~(ii), squaring the per-feature bounds and summing over all PPV
coordinates yields
\begin{equation}
    \|\Phi_s(\mathbf{y})
    -\Phi_s(\mathbf{x})\|_2^2
    \le
    \frac{s^2}{16}
    \left(
        \sum_{k=1}^{K}
        \frac{G_k^2}{N'_k}
    \right)
    \|\mathbf{y}-\mathbf{x}\|_2^2.
\end{equation}
Taking square roots proves the global bound.
\end{proof}

Equation \eqref{eq:lipschitz} shows that $\Phi_s$ is globally Lipschitz, with $C_s$ a
Lipschitz constant for the representation. The important point is the
structure of the corresponding per-feature bound:
\begin{equation*}
    \text{sensitivity bound}
    \;\sim\;
    \frac{
        \text{threshold sharpness}
        \times
        \text{kernel amplification}
    }{
        \sqrt{\text{number of pooled responses}}
    }.
\end{equation*}

The bound separates the contributions of convolution and pooling: greater spectral gain increases the upper bound, whereas averaging more responses reduces it, holding the other quantities fixed. The latter is directly related to the input length. For stride one and no padding,
\begin{equation}
N'_k = N-(l_k-1)d^{(k)},
\end{equation}
whereas length-preserving padding gives $N'_k=N$. Thus, for fixed kernel parameters, longer inputs generally lead to more responses being averaged and consequently to a smaller upper bound on sensitivity.

\subsection*{Spectral scaling for unit-energy ROCKET kernels}
\label{app:rocket-scaling}

We now specialise the stability bound to the unit-energy kernels used in
Section~\ref{subsec:experimental_protocol}, relating the spectral gain $G_k$
to kernel length and examining the resulting scaling of $C_s$.

Let $\mathbf{v}^{(k)}$ be a mean-centred standard Gaussian kernel of length
$l\geq2$, and let $\mathbf{w}^{(k)}=\mathbf{v}^{(k)}/\|\mathbf{v}^{(k)}\|_2$.
Its frequency response is
\begin{equation}
    W_k(\omega)
    =
    \sum_{j=1}^{l}
    w^{(k)}_j\,e^{-{\rm i}\omega (j-1)},
\end{equation}
so that $G_k=\sup_\omega|W_k(\omega)|$ as in \eqref{eq:spectral_gain}.
For the corresponding unnormalised spectral gain $H_k$, standard estimates
for Gaussian trigonometric polynomials give
\begin{equation}
    \mathbb{E}[H_k^2]=O(l\log l).
    \label{eq:gaussian_spectral}
\end{equation}
The mean-centred Gaussian is isotropic in the $(l-1)$-dimensional subspace
of zero-sum vectors. Its radius $R_k=\|\mathbf{v}^{(k)}\|_2$ and direction
$\mathbf{w}^{(k)}$ are independent, with $\mathbb{E}[R_k^2]=l-1$.
Since $H_k=R_kG_k$, it follows that
\begin{equation}
    \mathbb{E}[G_k^2]
    =\frac{\mathbb{E}[H_k^2]}{l-1}
    =O(\log l).
    \label{eq:unit_energy_spectral}
\end{equation}

If $K$ independent kernels have common length $l$ and common fixed output
length $N'$, then
\begin{equation}
    C_s^2
    =
    \frac{s^2}{16N'}
    \sum_{k=1}^{K}G_k^2.
\end{equation}
Taking expectations gives
\begin{align}
    \mathbb{E}[C_s^2]
    &=\frac{s^2}{16N'}\sum_{k=1}^{K}\mathbb{E}[G_k^2]
    \nonumber\\
    &=O\left(\frac{s^2K\log l}{N'}\right).
\end{align}
Therefore,
\begin{equation}
    \sqrt{\mathbb{E}[C_s^2]}
    =O\left(s\sqrt{\frac{K\log l}{N'}}\right).
    \label{eq:rocket_scaling}
\end{equation}
The factor $\sqrt{K}$ is an accumulation over coordinates of the
representation, not an increase in the sensitivity of any individual feature.
Unit-energy normalisation removes the additional factor $l$ from the
expected squared spectral-gain bound, weakening its explicit dependence on
kernel length. Length can still affect the number of pooled responses $N'_k$,
particularly without padding; the general bound in
\eqref{eq:stability_constant} covers differing output lengths.
This result concerns smooth PPV at fixed $s$ and does not give a uniform
Lipschitz guarantee for hard PPV as $s\to\infty$.

\section{Proofs of the concentration results}
\label{app:concentration}

\subsection{Proof of Theorem~\ref{thm:pointwise_concentration}}
\label{app:proof_concentration}

Fix the two signals $\mathbf{x}$ and $\mathbf{x}'$, and write
\begin{equation}
    g_k
    =
    \left(
        \phi_{\theta_k}(\mathbf{x})-\phi_{\theta_k}(\mathbf{x}')
    \right)^2,
    \qquad k=1,\ldots,K.
\end{equation}

\emph{Boundedness.} Each PPV feature is an average of indicator functions and
therefore satisfies $\phi_{\theta_k}(\mathbf{x})\in[0,1]$. Hence
$\phi_{\theta_k}(\mathbf{x})-\phi_{\theta_k}(\mathbf{x}')\in[-1,1]$, and
consequently
\begin{equation}
    g_k \in[0,1].
    \label{eq:g_bounded}
\end{equation}

\emph{Independence.} The parameters $\theta_1,\ldots,\theta_K$ are sampled
i.i.d.\ from $P$, and $g_k$ is a fixed measurable function of $\theta_k$ alone,
the two signals being held fixed. The variables $g_1,\ldots,g_K$ are therefore
i.i.d., with common mean
\begin{equation}
    \mathbb{E}[g_k]
    =
    \rho_P(\mathbf{x},\mathbf{x}'),
\end{equation}
by the definition~\eqref{eq:population_distance} of the population separation.
In particular, $\rho_K(\mathbf{x},\mathbf{x}')$ of~\eqref{eq:empirical_distance}
is an unbiased estimator of $\rho_P(\mathbf{x},\mathbf{x}')$.

\emph{Concentration.} Hoeffding's inequality for independent variables $g_k$
taking values in $[a_k,b_k]$ gives, for every $\varepsilon>0$,
\begin{equation}
    \mathbb{P}
    \left(
        \left|
            \frac{1}{K}\sum_{k=1}^{K}g_k
            -
            \mathbb{E}\!\left[\frac{1}{K}\sum_{k=1}^{K}g_k\right]
        \right|
        \ge\varepsilon
    \right)
    \le
    2\exp\left(
        -\frac{2K^2\varepsilon^2}{\sum_{k=1}^{K}(b_k-a_k)^2}
    \right).
\end{equation}
By \eqref{eq:g_bounded} we may take $a_k=0$ and $b_k=1$ for every $k$, so that
$\sum_{k=1}^{K}(b_k-a_k)^2=K$. Substituting yields
\begin{equation}
    \mathbb{P}
    \left(
        \left|
            \rho_K(\mathbf{x},\mathbf{x}')-\rho_P(\mathbf{x},\mathbf{x}')
        \right|
        \ge\varepsilon
    \right)
    \le
    2e^{-2K\varepsilon^2}. \notag
\end{equation}

\subsection{Proof of Corollary~\ref{cor:uniform_band}}
\label{app:proof_uniform}

Let $\mathcal{P}=\{(i,i'):1\le i<i'\le M\}$, so that
$|\mathcal{P}|=\binom{M}{2}$. For each pair, define the event
\begin{equation}
    A_{i,i'}
    =
    \left\{
        \left|
            \rho_K(\mathbf{x}^{(i)},\mathbf{x}^{(i')})
            -
            \rho_P(\mathbf{x}^{(i)},\mathbf{x}^{(i')})
        \right|
        >\varepsilon
    \right\}.
\end{equation}
The same $K$ kernels are used for every pair, so these events are not
independent. A union bound requires no independence, however, and
Theorem~\ref{thm:pointwise_concentration} applies to each pair separately,
giving
\begin{equation}
    \mathbb{P}
    \left(
        \bigcup_{(i,i')\in\mathcal{P}}A_{i,i'}
    \right)
    \le
    \sum_{(i,i')\in\mathcal{P}}
    \mathbb{P}\left(A_{i,i'}\right)
    \le
    2\binom{M}{2}e^{-2K\varepsilon^2}.
\end{equation}
Setting the right-hand side equal to $\delta$ and solving for $\varepsilon$
gives
\begin{equation}
    \varepsilon
    =
    \sqrt{
        \frac{\log\!\left(2\binom{M}{2}\big/\delta\right)}{2K}
    }.
\end{equation}
With this choice, the complement of
$\bigcup_{(i,i')\in\mathcal{P}}A_{i,i'}$ has probability at least $1-\delta$,
and on that event the stated maximum is at most $\varepsilon$. \hfill$\square$

\end{document}